\documentclass[letterpaper, 10 pt, conference]{ieeeconf}  

\IEEEoverridecommandlockouts                              

\usepackage{graphics} 
\usepackage{epsfig} 
\usepackage{times} 
\usepackage{amssymb}  
\usepackage{bm}
\usepackage{xcolor}
\usepackage{cite}

\usepackage{ulem} 
\usepackage{hyperref}

\usepackage{amsthm}
\usepackage{float}
\usepackage{booktabs}
\usepackage{multirow}
\usepackage{makecell}
\usepackage{enumerate}
\usepackage{leftindex}
\usepackage{mathtools}
\usepackage{mathrsfs}
\usepackage[utf8]{inputenc}
\usepackage[caption=false,font=footnotesize]{subfig}
\usepackage{graphicx}
\usepackage{pifont}
\usepackage{soul}
\usepackage[ruled,linesnumbered]{algorithm2e}  
\usepackage{bbding}

\newcounter{RNum}
\renewcommand{\theRNum}{\arabic{RNum}}

\theoremstyle{plain}
\newtheorem{assumption}{Assumption}
\newtheorem{theorem}{Theorem}
\theoremstyle{definition}

\newcommand{\Remark}{\noindent\textbf{Remark}~\refstepcounter{RNum}\textbf{\theRNum}: }
\newcommand{\NoOne}[1]{\textcolor{red}{#1}}
\newcommand{\NoTwo}[1]{\textcolor{green}{#1}}
\newcommand{\NoThree}[1]{\textcolor{blue}{#1}}

\newcommand{\re}{\mathbb{R}}
\newcommand{\qm}{\mbox{Qmod}}

\soulregister\NoOne7
\soulregister\NoTwo7
\soulregister\NoThree7
\soulregister\Remark7

\title{\LARGE \bf
Geometry-Aware Safety-Critical Local Reactive Controller for Robot Navigation in Unknown and Cluttered Environments  
}

\author{Yulin Li$^{1,*}$, Xindong Tang$^{2,*}$, Kai Chen$^{3}$, Chunxin Zheng$^{3}$, Haichao Liu$^{3}$, and Jun Ma$^{1}$ 
 \thanks{$^{*}$indicates equal contribution.}%
\thanks{$^{1}$Yulin Li and Jun Ma are with the Department of Electronic and Computer Engineering, The Hong Kong University of Science and Technology, Hong Kong SAR, China (e-mail: yline@connect.ust.hk; jun.ma@ust.hk).
}%
\thanks{$^{2}$Xindong Tang is with the Department of Mathematics, Hong Kong Baptist University, Hong Kong SAR, China (e-mail: xdtang@hkbu.edu.hk).
}%
\thanks{$^{3}$Kai Chen, Chunxin Zheng and Haichao Liu are with the Robotics and Autonomous Systems Thrust, The Hong Kong University of Science and Technology (Guangzhou), Guangzhou, China (e-mail: kchen916@connect.hkust-gz.edu.cn; czheng739@connect.hkust-gz.edu.cn; hliu369@connect.hkust-gz.edu.cn).}
}

\begin{document}

\maketitle
\thispagestyle{empty}
\pagestyle{empty}


\begin{abstract}

This work proposes a safety-critical local reactive controller that enables the robot to navigate in unknown and cluttered environments.  
In particular, the trajectory tracking task is formulated as a constrained polynomial optimization problem. 
Then, safety constraints are imposed on the control variables invoking the notion of polynomial positivity certificates in conjunction with their Sum-of-Squares (SOS) approximation, thereby confining the robot motion inside the locally extracted convex free region. 
It is noteworthy that, in the process of devising the proposed safety constraints, the geometry of the robot can be approximated using any shape that can be characterized with a set of polynomial functions.
The optimization problem is further convexified into a semidefinite program (SDP) leveraging truncated multi-sequences (tms) and moment relaxation, which favorably facilitates the effective use of off-the-shelf conic programming solvers, such that real-time performance is attainable. Various robot navigation tasks are investigated to demonstrate the effectiveness of the proposed approach in terms of safety and tracking performance.
\end{abstract}

\section{Introduction} \label{sec:intro}
As one of the core functionalities, safety in motion has emerged as an indispensable and critical aspect of robotic autonomy, driven by the increasing need for robots to interact seamlessly with their surrounding environments. Safety is generally ensured by planning and executing a feasible trajectory globally in the free space, assuming complete knowledge of the environment \cite{choset2005principles,latombe2012robot}. Unfortunately, there could be situations where the environment is partially or fully unknown. Under such circumstances, it is typically desirable to design a highly reactive local planner or controller to facilitate swift and accurate responses to unforeseen environmental changes, thereby adapting the global path in real time to ensure collision-free maneuvers.


\begin{figure}[t]	
	\centering
	\includegraphics[trim=2cm 4.5cm 2cm 0cm, clip,width=1\linewidth]{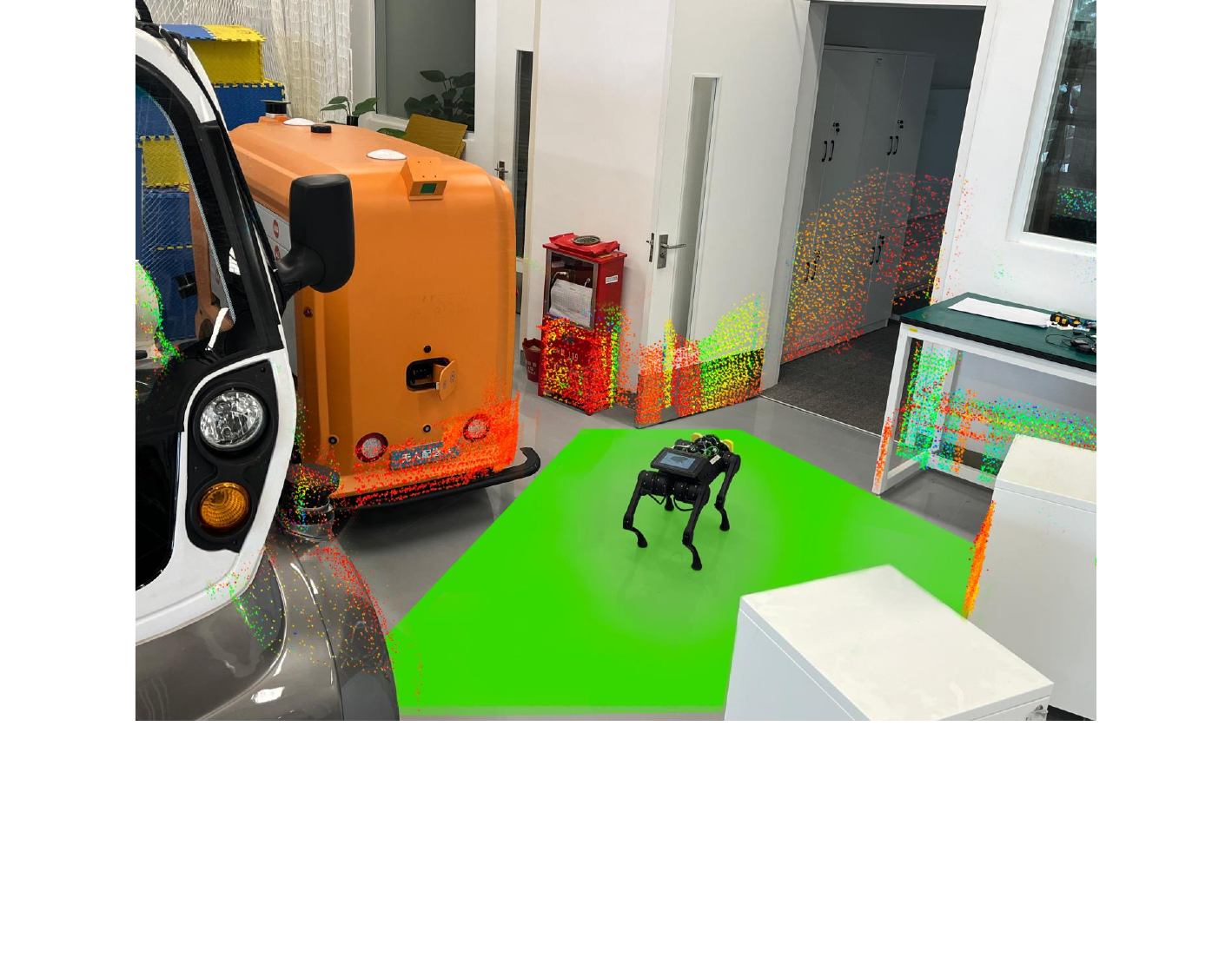}
	\setlength{\abovecaptionskip}{-15pt} 
	\caption
	{Overview of the proposed methodology. The proposed reactive controller aims to minimize the tracking error while constraining the robot with specific geometry inside the free region (shown in green) by solving a constrained polynomial optimization problem with safety-related polynomial no-negativity constraints.}
	\label{fig:illustration_real_platform}
\end{figure}

 Serving as one of the underpinning basis, optimization-based methods have garnered significant interest among researchers for catering to such problems. Certainly, a natural and straightforward choice is to incorporate safety constraints by enforcing a distance margin between the robot and obstacles. As an evitably encountered challenge, the distance calculation necessitates the knowledge of obstacle geometry, and maintaining this distance for the robot with each surrounding obstacle induces heavy burdens to the requisite computation efficiency, particularly in cluttered environments. An alternative employs distance field representations, such as Euclidean Signed Distance Field (ESDF) \cite{9145591,9196996}, to encode obstacle information, which can be further modeled as soft constraints and integrated into the cost function\cite{9422918,9145591}. 
 A novel local reactive controller has recently been proposed in \cite{mattamala2022efficient} that designs acceleration fields combining GDF and SDF as Riemannian Motion Policies (RMPs)~\cite{wingo2020adaptively}, and safe goal reaching behaviors in unknown environments are achieved for the quadruped by solving a least square problem. However,
 the computation of the local vector field representation can be a time-intensive task, which may hinder the real-time applicability of these approaches. 
On the other hand, approximating free space with convex shapes\cite{7839930, deits2015computing} and confining robot motion within the extracted convex area is far more computationally efficient in obstacle-dense environments than distance-based methods. Yet in most of the scenarios, determining whether one shape is entirely contained within another is burdened by intricacy, posing significant challenges for incorporating such conditions into the optimization problem formulation. Consequently, polynomial trajectories of point-robot models are typically employed to simplify the situation\cite{7839930}. Recent advances in polynomial positivity certificates \cite{putinar1993positive} and Sums-of-Squares (SOS) programming \cite{sos1} have shown great potential in reasoning about the containment relationship between two general shapes \cite{dai2023certified}. In \cite{7138978}, polynomial positivity certificates are utilized to ensure that the entire polynomial trajectory remains within the free regions. However, these methods are limited to the application on two stationary objects and cannot be directly applied to ensure robot motion safety involved with control. 
In this sense, Control Barrier Function (CBF)-based methods are brought to attention in designing safety-critical controllers \cite{8796030}. Nonetheless, CBF must be continuously differentiable with respect to state variables, potentially leading to over-approximation of collision geometry for explicit distance expressions and conservative behaviors~\cite{ferraguti2020control}. 
In \cite{9812334}, CBF-based safety constraints are derived based on the dual form of obstacle avoidance between two polytopes and incorporated in the Model Predictive Control (MPC)  framework\cite{rosolia2020multi}.
Although this approach demonstrates effective maneuvers in environments with numerous polytopic obstacles, the inequality scaling arising from the derivation of a valid CBF function for polytopic geometry restricts the feasible set of control variables. Besides, specific system models and robot geometries need to be taken into appropriate consideration. 
With all the above descriptions as a backdrop, a real-time non-conservative obstacle avoidance algorithm for general robotic systems that considers tight-fitted robot geometry remains an open and challenging problem.

To address the aforementioned limitations, this work presents a local reactive controller for robots to safely navigate in unknown and cluttered environments. 
We propose a constrained polynomial optimization problem to obtain the optimal twist command for tracking the reference path within the feasible set of control variables. Subsequently, we incorporate the geometry-aware safety constraints into the proposed polynomial optimization framework. We show that the robot configuration and geometry after applying the control can be expressed as polynomial functions of twist \cite{lynch2017modern}, whereupon the safety constraints are derived using certificates of nonnegative polynomials \cite{nie2013certifying}. To render the proposed optimization problem computationally tractable, we first approximate the cone of non-negative polynomials in the safety constraint using its quadratic module\cite{nie2014truncated}. Furthermore, we linearize and convexify the safety constraints with truncated multi-sequences (tms) and moment relaxation, and the original polynomial optimization problem with safety conditions is relaxed to a linear semidefinite program (SDP) that can be solved in real time.
In summary, the main contributions of our work are as follows:
 \begin{itemize}
    \item A geometry-aware local reactive controller for robot navigation tasks is presented, which is particularly well-suited for deployment in unknown and cluttered environments utilizing locally extracted free regions.
 
\item In the proposed constrained polynomial optimization framework, the feasible set of control variables and the objective function can be effectively modeled with polynomial functions. 
In this sense, safety constraints 
are suitably established for a wide range of robots with general geometries.


\item We further relax the proposed constrained polynomial optimization problem with nonnegativity constraints to an SDP leveraging 
SOS approximation and moment relaxation, which render it possible to generate reactive motion in real time.
    

\item The proposed controller is implemented in both numerical simulations and real-world experiments with comprehensive evaluations, and the results demonstrate the effectiveness and efficiency of the proposed controller in robot navigation tasks.

\end{itemize}

\section{Preliminaries}
\label{sec:preliminary}
\subsection{Twist Theory for Rigid Body Motion}\label{twist_preliminary}
Instead of considering a control strategy for a specific robot, we use twist $\mathcal{V} = (\omega, v) \in \re^{6}$ to describe rigid body motion as a displacement along a fixed screw axis $\mathcal{S} \in \re^6$ for unit time at a speed $\dot{\theta}\in \re$.

We denote zero vector in $\re^3$ as $\mathbf{0}_{\re^3}$, and if $\omega \neq \mathbf{0}_{\re^3}$, $\mathcal{S}=\mathcal{V} /\|\omega\|=(\hat{\omega}, \hat{v})$, which is simply $\mathcal{V}$ normalized by the length of the angular velocity vector. The velocity about the screw axis is $\dot{\theta}=\|\omega\|$, such that $\mathcal{S} \dot{\theta}=\mathcal{V}$. On the other hand, if no rotational motion is involved, namely $\omega=\mathbf{0}_{\re^3}$, 
the screw axis $\mathcal{S}$ becomes $\mathcal{V}$ normalized by the length of the linear velocity vector and $\dot{\theta}=\|v\|$. 

The homogeneous transformation $T \in SE(3)$ generated by moving with control command $\mathcal{V}$ for step time $\Delta t$ can be represented using the exponential coordinates, where

\begin{equation}
\label{eq:exponential_expression}
\begin{gathered}
\mathcal{V} \Delta t=S \dot{\theta} \Delta t=S \theta_{\Delta t}, \\
\end{gathered}  
\end{equation}
\begin{equation}
\begin{gathered}
    T=e^{[\mathcal{S}] \theta_{\Delta t}}=
\left[\begin{array}{cc}
e^{[\hat{\omega}] \theta_{\Delta t}} & G\left(\theta_{\Delta t}\right) \hat{v} \\
0 & 1
\end{array}\right],
\end{gathered}
\end{equation}

with
\begin{equation}
\label{eq:exponential_rotation}
 \begin{aligned}
e^{[\hat{\omega}] \theta_{\Delta t}} &=I+\sin (\theta_{\Delta t})[\hat{\omega}]+(1-\cos (\theta_{\Delta t}))[\hat{\omega}]^2 \\
& =R(\hat{\omega}, \theta_{\Delta t}) \in SO(3),
\end{aligned}   
\end{equation}
\begin{equation}
\label{eq:exponential_translation}
   \begin{aligned}
G\left(\theta_{\Delta} t\right)\hat{v} & =I \theta_{\Delta t}\hat{v}+\left(1-\cos (\theta_{\Delta t})\right)[\hat{\omega}]\hat{v}\\& \,\,\,\,\,\,+\left(\theta_{\Delta t}-\sin (\theta_{\Delta t})\right)[\hat{\omega}]^{2}\hat{v}  \\&= p(\hat{w},\hat{v},\theta_{\Delta t}) \in \re^3.
\end{aligned} 
\end{equation}
Note that $[\hat{\omega}] \in so(3)$ is the skew-symmetric representation of 
$\hat{\omega}$.

\subsection{Polynomial Functions and Sums-of-Squares}
Let $x = (x_1,x_2\ldots,x_n)$ be a vector in $\re^n$, and denote $\re[x]$ as the space of all real-coefficient polynomial functions in $x$. For a given order $d$, $\re[x]_d$ denotes the set of all polynomials with degrees no greater than $d$.

Given the tuple $\alpha:=(\alpha_1,\alpha_2\dots,\alpha_n)$ of nonnegative integers,
$x^{\alpha}:=x_1^{\alpha_1}x_2^{\alpha_2}\dots x_n^{\alpha_n}$ is called a \textbf{\textit{monomial}} in $x$ with degree $d:=\alpha_1+\alpha_2\dots+\alpha_n$. We define $[x]_d$ as the vector of all monomials in $x$ whose degrees are not greater than $d$. Then, for a multivariate polynomial function $p(x)\in \re[x]$, we can uniquely express it with a linear combination of its monomials vector:
$$p(x) = {\rm coef}(p)^{\top} [x]_{\operatorname{\operatorname{\deg}(p)}},$$
 where ${\rm coef}(p)$ is the vector of coefficients of $p$ corresponding to each monomial in $[x]_d$.

Next, we introduce the concept of a special class of polynomials called \textit{\textbf{Sums-of-Squares}} (SOS) polynomials. A polynomial $\sigma(x)\in \re[x]$ is an SOS if 
$\sigma = p_1^2+p_2^2\ldots+ p_r^2$
for some $p_1,p_2\ldots,p_{r}\in \re[x]$.
Denote the set of SOS polynomials as $\Sigma[x]$.
Then $\sigma \in \Sigma\cap \re[x]_{2d}$ if and only if it can be written in a quadratic form:
$$\sigma = [x]_d^{\top} X [x]_d,$$
where $X\in \mathbf{S}^d_+$ is a symmetric positive semidefinite matrix. 

\subsection{Nonnegative Polynomials and SOS Approximation}
Nonnegativity polynomials and their SOS approximation are useful for establishing the containment relationship of two general geometries \cite{dai2023certified}, which serves as the essential basis for deriving the geometry-aware safety constraints for the robot.

Consider the following set defined with $m$ polynomial inequalities in $x$:
\[ K:= \{ x\in\re^n: f_1(x)\ge0,f_2(x)\ge0,\ldots, f_m(x)\ge0 \},\]
where $f_i(x) \in \re[x]$  for all $i=1,2,\ldots, m$. Let $\mathcal{P}(K)$ be the cone of nonnegative polynomials on $K$:
\[ \mathcal{P}(K) = \{ p\in \re[x]: p(x)\ge0 \ \forall x\in K \}. \]
To render $\mathcal{P}(K)$ computational tractable, we introduce the \textit{\textbf{quadratic module}} of $f$, which is defined as:
\[ \qm[f] :=  \{ \sigma_0 + f_1 \sigma_1 + \ldots + f_m\sigma_m: \sigma_0,\sigma_1,\ldots,\sigma_m \in \Sigma[x]\}. \]
Then $\qm[f]\subseteq \mathcal{P}(K)$.
However, the converse is not true in general.
In the following, we review the \textit{\textbf{Putinar Positivestellensatz}}, which provides theoretical guarantee on how close if we use $\qm[f]$ to approximate a polynomial in $\mathcal{P}(K)$.


\begin{theorem}\textup{\cite{putinar1993positive}}
\label{tm:putinar}
    Suppose {\rm $\qm[f]$} is \textbf{archimedean}\footnote{We say $\qm[f]$ is {\it archimedean} if there exists $\rho\in \qm[f]$ such that the set given by $\rho(f)\ge 0$ is bounded.}.
    If $p(x)>0$ for all $x\in K$, then $p\in${\rm $\qm[f]$}.
\end{theorem}

\Remark\label{remark:archimedea}{Archimedeanness is a property that is slightly stronger than set compactness. If $K$ is compact, one can always modify the polynomial tuple $f$ by adding a redundant ball constraint such that the archimedeanness holds.}



\subsection{Truncated Multi-Sequences and Moment Problems}
Let $J(u)$ be a polynomial in $u\in \re^n$, 
consider the following constrained polynomial optimization problem: 
\begin{equation}\label{eq:pop} 
\min_{u} \  J(u)\quad  \mbox{s.t.} \quad  u\in G. 
\end{equation}
It aims to minimize $J(u)$ over a feasible set $G:=\{u: g_i(u)\ge0\ \forall i=1,2,\ldots,m\}$ defined with a series of polynomial inequalities in $u$. One should recognize that (\ref{eq:pop}) is inherently a nonlinear nonconvex optimization problem that is hard to solve, we now introduce the basis of its semidefinite relaxation scheme.

We call $y$ a \textit{\textbf{truncated multi-sequence}} (tms) if it is labeled by the tuple of its monomial powers $\alpha = (\alpha_1, \alpha_2, \dots, \alpha_n)$. 
There exists a natural mapping that maps every monomial $x^{\alpha}$ up to degree $d$ to $y_{\alpha}$,
which generates the \textit{\textbf{Riesz functional}} of a polynomial function $p(x)$:
\[ \mathcal{L}_y(p) = \sum_{\alpha} p_{\alpha}y_{\alpha},\ \forall p := \sum_{\alpha} p_{\alpha}x^{\alpha}\in\re[x]_d.\]

Note that tms can be applied to formulate moment relaxations of (\ref{eq:pop}).
Let $\lceil z \rceil$ be the ceiling of the real scalar $z$.
For the given tms $y$ and $j = 1,2,\dots,m$
and $\ell \ge \ell_0:= \lceil \max\{\operatorname{deg}(g_i)/2, \operatorname{deg}(J)/2\}\rceil $,
the $\ell$th \textit{\textbf{localizing matrix}} of $g_j$ generated by $y$ is
\[L^{(\ell)}_{g_j}[y] : = \mathcal{L}_y(g_j\cdot [x]_{ \ell_{j} }[x]_{\ell_{j}}^T),\]
where $\ell_j:=\lceil (2\ell-\operatorname{\deg}{g_j})/2 \rceil$.
Particularly, $M_\ell[y] = \mathcal{L}_y([x]_\ell[x]_\ell^T)$ is called the $\ell$th\textit{\textbf{ moment matrix}} of $y$. Both the moment and localizing matrices are obtained by applying $\mathcal{L}_y$ to each entry of the polynomial matrix.
Then, the $\ell$th order \textit{\textbf{moment relaxation}} for (\ref{eq:pop}) is:
\begin{equation}\label{eq:moment_relax}
\begin{array}{cl}
\displaystyle \min_{y} & \mathcal{L}_{y}(J) \\
\mbox{s.t.} & y_{\bf0} =1,\ y\in\mathscr{S}_{\ell}(g),
\end{array}
\end{equation}
where $y_{\bf0}:=\mathcal{L}_y(1)$, and the \textit\textbf{{moment cone}} $\mathscr{S}$ is defined as:
\[\mathscr{S}_{\ell}(g) := \left\{ y \left|
\begin{array}{c}

M_{\ell}[y]\succeq 0, \ L^{(\ell)}_{g_i}[y]\succeq 0, \forall\,
i =1,2,\dots,m
\end{array}
\right. \right\}. \]
Notably, (\ref{eq:moment_relax}) is an SDP, the objective function $J$ is linearized with the tms $y$, and $\mathscr{S}$ is the intersection of the cone of positive semidefinite matrices. 
If $\qm[g]$ is archimedean, then the optimum of (\ref{eq:moment_relax}) monotonically converges to the global optimum of (\ref{eq:pop}) as $\ell$ increases\cite{lasserre2001global}.
\begin{figure}[t]	
	\centering
        \vspace{3pt}
	\includegraphics[trim=1cm 0.8cm 0 0.90cm, clip,width=0.9\linewidth]{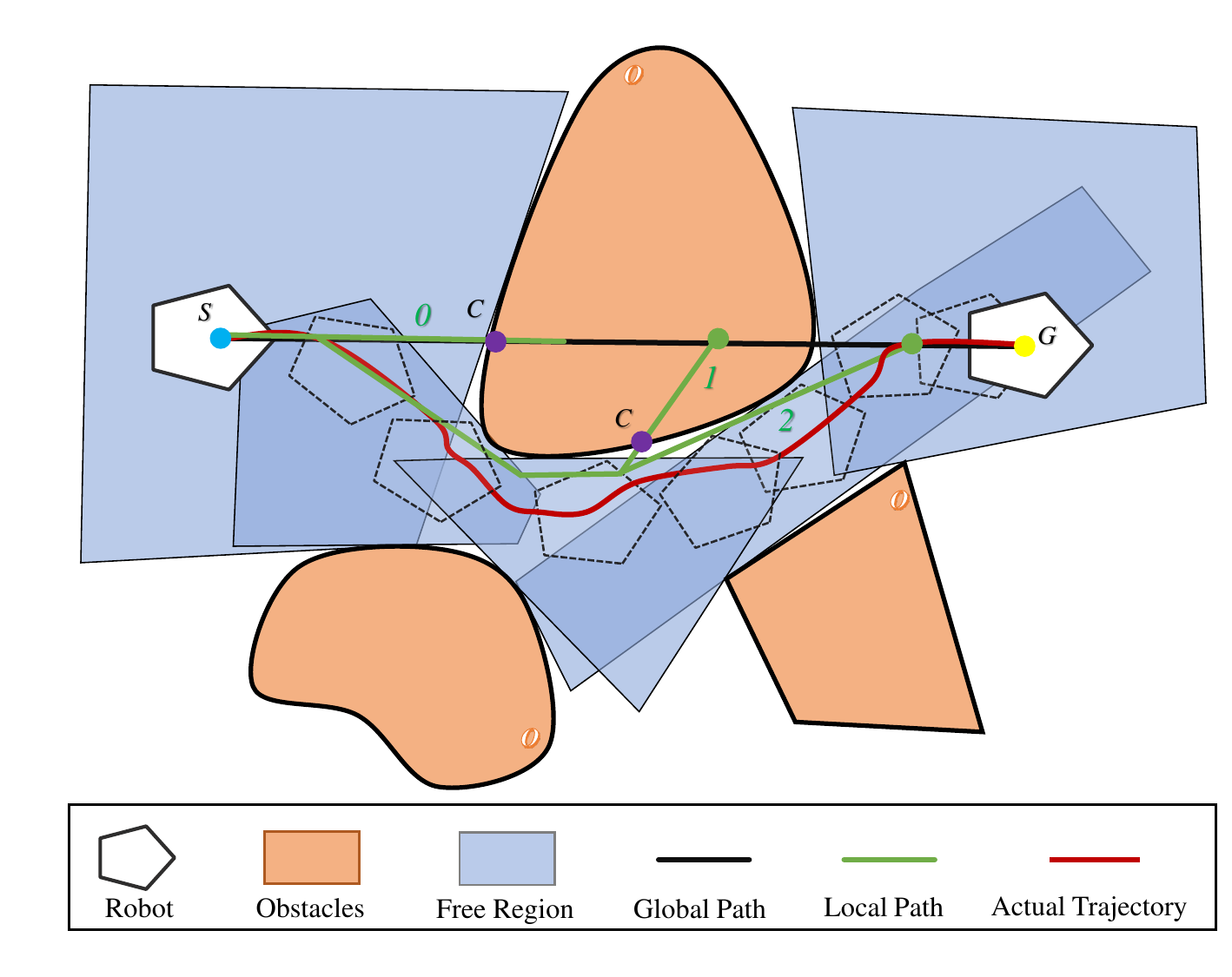}
	\setlength{\abovecaptionskip}{-0pt} 
	\caption
	{The robot is traversing in an unknown environments from the start configuration \textit{S} to goal configuration \textit{G} along the local reference extracted from global path. The local reference is re-generated once potential collision is detected on the path (at those purple points).}
	\label{fig:overview}
\end{figure}

\section{Methodology}
\subsection{Control Problem Formulation}
 Observed from Sec. \ref{twist_preliminary}, if we fix $\theta_{\Delta t}$ (the distance traveled along the screw axis), the homogeneous transformation $T:=(R,p)$ can be represented as polynomial functions of the screw axis $(\hat{\omega}, \hat{v})$ up to degree 3 (in the term $[\hat{\omega}]^2\hat{v}$). 
 On the other hand, rigid body motion can be unified via the exponential representation with $\hat{\omega}$ being zero (pure translation) or a unit vector (rotation involved motion). 
 We now introduce our control problem: given $\leftindex[V]^bT_r:=(\leftindex[V]^bR_r,\leftindex[V]^bp_r)$ the reference configuration expressed in robot body frame $\{b\}$. We aim to optimize the body screw axis $\mathcal{S}$ within its feasible set $G$ for tracking the reference, which can be naturally modeled as the following polynomial optimization problem:
 \begin{equation}
\label{eq:poly_opt_control_problem}
\begin{aligned}
    &\min _{\hat{\omega}\in\re^3, \hat{v}\in\re^3}   J(\hat{\omega}, \hat{v}) 
\quad\text { s.t. } \quad(\hat{\omega}, \hat{v}) \in G.
\end{aligned}
\end{equation}
The cost function $J$ and feasible set $G$ are defined with polynomial functions in ($\hat{\omega},\hat{v}$). Denote the column-wise vectorized rotation matrix as $\operatorname{vec}(R)\in \re^9$, the cost function is designed to minimize both the position error $e_p:=\leftindex[V]^bp(\hat{w}, \hat{v})-\leftindex[V]^bp_r$ and orientation error $e_R:=\operatorname{vec}(\leftindex[V]^bR(\hat{w}, \hat{v}))-\operatorname{vec}(\leftindex[V]^bR_r)$:
\begin{equation}
 J = J_p+J_R 
= e_p^{\top} Q_pe_p 
+ e_R^{\top} Q_Re_R.
\end{equation}
In the above, $Q_p\in\mathbf{S}^3_+$ and $Q_R\in\mathbf{S}^9_+$ are the weighting matrix for tracking pose and orientation error respectively.
In the tracking task described above, we impose the following constraints on the decision variables: 
\begin{equation}\label{eq:G}
        G=\left\{ u = (\hat{\omega},\hat{v}) \in\re^6 \left|\begin{array}{c}(\hat{\omega}^\top\hat{\omega})\cdot(\hat{\omega}^\top\hat{\omega}-1) = 0,\\
        v^2_{limit}-\hat{v}^\top\hat{v}\ge0
        \end{array}
        \right.\right\}.
\end{equation}
In (\ref{eq:G}), the equality states that $\hat{\omega}$ must be zero or a unit vector to keep consistent with the definition of screw axis, while the inequality limits the tracking velocity $\hat{v}$.

\Remark{The proposed polynomial optimization framework enables the inclusion of adding any polynomial functions in the feasible set $G$ and cost function $J$.}
\subsection{Safety Constraints}
As shown in Fig. \ref{fig:overview}, our controlled robot is traversing in an obstacle-dense environment, and safety is attained by confining the robot motion inside the surrounding convex free region. In this section, we derive the safety constraints into the proposed polynomial optimization problem (\ref{eq:poly_opt_control_problem}). 

At the time $t$, geometries of the robot $A$ and free region $B$ in body frame are defined with sets of polynomial inequalities as:
\begin{equation}
\begin{gathered}
\label{def:robot_obstacles}
    \leftindex[V]^bA^t:=\{x\in\re^n: \leftindex[V]^bf^t_{A_1}(x)\geq 0, \ldots, \leftindex[V]^bf^t_{A_a}(x)\ge0\},\\
    \leftindex[V]^bB^t:=\{x\in\re^n: \leftindex[V]^bf^t_{B_1}(x)\geq 0,\ldots, \leftindex[V]^bf^t_{B_b}(x)\ge0\}.
\end{gathered}
\end{equation}
Typically, $n$ is $2$ or $3$, corresponding to two or three dimensional Euclidean space respectively.
In this context, we mainly consider the case that the free region $B$ is a polytope whose representation consists of linear functions in $x$. However, the semialgebraic sets given by (\ref{def:robot_obstacles}) can represent the majority of convex shapes and some nonconvex shapes, e.g., double ring.

Denote $\leftindex[V]^bA^{t+\Delta t}$ the representation of robot geometry in $\{b\}$ after applying the current twist $\mathcal{V}_t$ for one time step $\Delta t$. $\mathcal{V}_t$ is said to be a safe control if $\leftindex[V]^bA^{t+\Delta t} \subset \leftindex[V]^bB^t $, which can also be interpreted as: 
$$
\leftindex[V]^bf^t_{B_i}(x) \in \mathcal{P}(\leftindex[V]^bA^{t+\Delta t})\quad \forall\, i=1,2,\ldots, b,
$$
where 
$\mathcal{P}(\leftindex[V]^bA^{t+\Delta t})$ is the cone of nonnegative polynomials on $\leftindex[V]^bA^{t+\Delta t}$.

\begin{figure}[h]	
	\centering
	\includegraphics[trim=1.5cm 1.5cm 1.6cm 1cm, clip,width=0.8\linewidth]{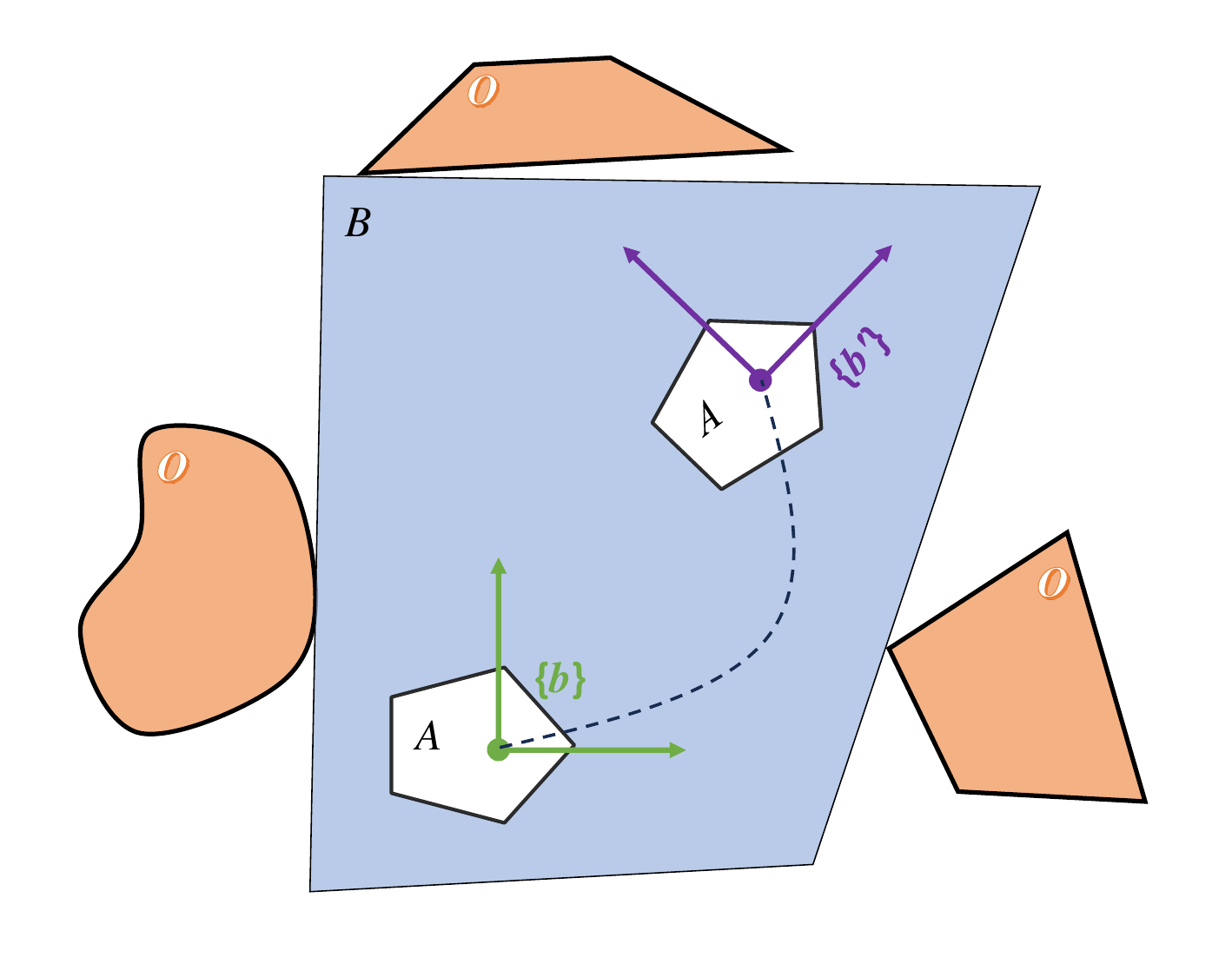}
	\setlength{\abovecaptionskip}{-0pt} 
	\caption
	{Illustration of safety conditions. Under the current twist command, the robot $A$ moves from $\{b\}$ to $\{b^\prime\}$ while staying in the extracted free region $B$.}
	\label{fig:illu_frame}
\end{figure}

 However, one should recognize that $\leftindex[V]^bf^{t+\Delta t}_{A}$ is dependent on the decision variables ($\hat{\omega},\hat{v}$), which renders it difficult to describe the nonnegative polynomial cone $\mathcal{P}(\leftindex[V]^bA^{t+\Delta t})$. To remove the dependency, we utilize the relativity of rigid body motion and instead evaluate the safety constraint from $\{b^\prime\}$. As indicated in Fig. \ref{fig:illu_frame}, the representation of robot geometry will not change with time in body frame, namely $\leftindex[V]^bA^{t} = \leftindex[V]^{b^\prime}A^{t+\Delta t}$. Incorporated with such safety constraints, the optimization problem (\ref{eq:poly_opt_control_problem}) now becomes:
\begin{equation}
\label{eq:add_safety_condition}
\begin{array}{cl}
\displaystyle   \min _{\hat{\omega}\in\re^3, \hat{v}\in\re^3}   &J(\hat{\omega}, \hat{v}) 
\\ \text {s.t.}  &\leftindex[V]^{b^\prime}f^t_{B_i}(x) \in \mathcal{P}(\leftindex[V]^{b^{\prime}}A^{t+\Delta t}) \quad\forall\, i = 1,2,\ldots,b\\
&(\hat{\omega}, \hat{v}) \in G. 
\end{array}
\end{equation}
In the safety constraints, we express $\leftindex[V]^{b^\prime}f^t_{B_i}(x)$ as the monomials vector multiplied by its coefficients, i.e., $\leftindex[V]^{b^\prime}f^t_{B_i}(x) := (\leftindex[V]^{b^\prime}c^t_{B_i}(\hat{w},\hat{v}))^\top[x]_{\operatorname{deg}(\leftindex[V]^{b^\prime}f^t_{B_i})}$, where the coefficients vector $\leftindex[V]^{b^\prime}c^t_{B_i}(\hat{w},\hat{v})$ consists of polynomial functions in the control variables $(\hat{w},\hat{v})$. 
\subsection{SDP Relaxation}
\subsubsection{Approximation of Nonnegative Polynomials}
First, we approximate the cone of nonnegative polynomials on $\leftindex[V]^bA^{t+\Delta t}$ with its quadratic module at a given order for describing the safety constraints explicitly.

Consider a relaxation order $k$ such that $2k\ge \max\{\operatorname{deg}(\leftindex[V]^{b\prime}f^{t+\Delta t}_{A}),\operatorname{deg}(\leftindex[V]^{b^\prime}f^t_{B}(x)\}$. Then the $k$th order SOS relaxation of (\ref{eq:add_safety_condition}) is:
\begin{equation}
\label{eq:add_safety_condition_qm}
\begin{array}{cl}
\displaystyle   \min _{\hat{\omega}, \hat{v}, \sigma_{i,j}}   &J(\hat{\omega}, \hat{v}) 
\\ \text { s.t.}  &  \leftindex[V]^{b^\prime}c^t_{B_i}(\hat{w},\hat{v}) = {\rm coef}(\sigma_{i,0}+\sum_{j=1}^a\sigma_{i,j} \cdot\leftindex[V]^bf^{t+\Delta t}_{A_j}),\\
&\quad\quad\forall\,\,i = 1,2,\ldots,b\\
& (\hat{\omega}, \hat{v}) \in G,\quad \sigma_{i,0}\in\Sigma[x]\cap \re[x]_{2k}, \\
&\sigma_{i,j}\in\Sigma[x]\cap \re[x]_{2k - \operatorname{deg}(\leftindex[V]^bf^{t+\Delta t}_{A_j})},\\
&\quad\quad\forall\,\, i = 1,2,\ldots,b,\,\forall\,\,j=1,2,\dots,a.

\end{array}
\end{equation}
In (\ref{eq:add_safety_condition_qm}), we use the $2k$ truncation of $\qm[\leftindex[V]^{b\prime}f^{t+\Delta t}_{A}]$ to approximate $\mathcal{P}(\leftindex[V]^{b^{\prime}}A^{t+\Delta t})$ in (\ref{eq:add_safety_condition}), where every $\sigma_{i,j}$ is the SOS multiplier in $\qm$ with a certain degree. We construct equality constraints directly from the coefficient vectors to establish $\leftindex[V]^{b^\prime}f^t_{B_i}(x) \in \qm_{2k}(\leftindex[V]^{b^{\prime}}A^{t+\Delta t})$, which eliminates the explicit relationship with $x$.
\begin{assumption}
\label{ass:b_in_p}
    Suppose $(\hat{\omega}^*,\hat{v}^*)$ is the optimal control attained by (\ref{eq:add_safety_condition}), there exists a feasible linear velocity $\hat{v}^0$ such that $(\hat{\omega}^*,\hat{v}^0) \in G$ and
    $$
    (\leftindex[V]^{b^\prime}c^t_{B_i}(\hat{\omega}^*,\hat{v}^0))^\top[x]_{d_i}>0\quad \forall x\in \leftindex[V]^{b^{\prime}}A^{t+\Delta t},\,\forall i=1,2,\dots,b.
    $$
\end{assumption}
We propose the following theorem to state the asymptotic convergence for approximating  
(\ref{eq:add_safety_condition}) with (\ref{eq:add_safety_condition_qm}).
\begin{theorem}\label{tm:sos_asymp}
    For every relaxation order $k$, denote $J^*$ and $J_k$ the optimal values for (\ref{eq:add_safety_condition}) and (\ref{eq:add_safety_condition_qm}), respectively. The minimizer of (\ref{eq:add_safety_condition_qm}) is a \textbf{safe} control with $J_k\ge J^*$, and $J_k$ is monotonically decreasing as $k$ increases.
    Furthermore, if Assumption \ref{ass:b_in_p} holds with $G$ convex in $\hat{v}$ and assume that $\qm[\leftindex[V]^bf^t_{A}(x)]$ is archimedean, 
    then $J_k\to J^*$ as $k\to \infty$.
\end{theorem}
\noindent\textit{Proof.} As $\qm[\leftindex[V]^{b^\prime}f^{t+\Delta t}_{A}]_{2k} \subseteq \mathcal{P}(\leftindex[V]^{b^\prime}A^{t+\Delta t})$ for all $k$, every feasible solution of (\ref{eq:add_safety_condition_qm}) is also feasible for (\ref{eq:add_safety_condition}), and thus the solution for (\ref{eq:add_safety_condition_qm}) is safe for robot to navigate in the surrounding free region.


Recall that the free region $B$ is given as a polytope. For a fixed $\hat{\omega}^*$, $\leftindex[V]^{b^\prime}c^t_{B_i}(\hat{\omega}^*,\hat{v})$ is linear in $\hat{v}$.
Let $\hat{v}^{\epsilon} = (1-\epsilon)\hat{v}^*+\epsilon\hat{v}^0$. If Assumption \ref{ass:b_in_p} holds, then for any $0< \epsilon \le  1$, it is clear that $(\hat{\omega}^*,\hat{v}^{\epsilon})\in G$.
Moreover, for all  $x\in \leftindex[V]^{b^{\prime}}A^{t+\Delta t}$, we have
\[
   \begin{aligned}
    (\leftindex[V]^{b^\prime}c^t_{B_i}(\hat{\omega}^*,\hat{v}^{\epsilon})^\top[x]_{d_i} =   \,&\epsilon (\leftindex[V]^{b^\prime}c^t_{B_i}(\hat{\omega}^*,\hat{v}^0)^\top[x]_{d_i} \\ &+ (1-\epsilon)(\leftindex[V]^{b^\prime}c^t_{B_i}(\hat{\omega}^*,\hat{v}^*)^\top[x]_{d_i}\\
    \ge& \epsilon (\leftindex[V]^{b^\prime}c^t_{B_i}(\hat{\omega}^*,\hat{v}^0)^\top[x]_{d_i}>0.
\end{aligned} 
\]
Hence, from Theorem~\ref{tm:putinar}, there exists $k_{\epsilon}\in \mathbb{N}$ such that for all $k\ge k_{\epsilon}$, $(\hat{\omega}^*,v^{\epsilon})$ is feasible for (\ref{eq:add_safety_condition_qm}).
Since $J(\hat{\omega},\hat{v})$ is a polynomial function, and thus it is Lipschitz continuous on the compact set $G$, and we have
\[ J(\hat{\omega}^*,v^{\epsilon}) - J(\hat{\omega}^*,v^*)\le M \Vert v^{\epsilon} - v^*\Vert = \epsilon M\cdot \Vert v^0\Vert, \]
where $M$ is the Lipschitz constant of $J(u)$.
Therefore, $J_k-J^* < \epsilon M\cdot \Vert v^0\Vert$, and $\epsilon$ can be a value arbitrarily close to 0. So we have $J_k\to J^*$ as $k\to \infty$. \hfill{\qed}
 
\Remark{The set $\leftindex[V]^{b^{\prime}}A^{t+\Delta t}$ representing robot geometry is compact, hence the archimedeanness assumption in Theorem~\ref{tm:sos_asymp} can always be satisfied as discussed in Remark \ref{remark:archimedea}. Besides, from the geometric perspective, Assumption \ref{ass:b_in_p} states that if we fix $\hat{\omega}$ as the optimal $\hat{\omega}^*$, there exists a feasible $\hat{v}^0$, such that the robot will stay strictly in the interior of the free region $ \leftindex[V]^{b^\prime}B^t$} without hitting the boundary after applying $(\hat{\omega}^*,\hat{v}^0)$.

\subsubsection{Moment Relaxation and Linearization of Safety Constraints}
Since $\leftindex[V]^{b^\prime}c^t_{B_i}(\hat{\omega},\hat{v})$ is not linear in $u$, it is still difficult to solve (\ref{eq:add_safety_condition_qm}) directly.
Therefore, we consider the moment relaxation of (\ref{eq:add_safety_condition_qm}):
\begin{equation}
\label{eq:mom_relx}
\begin{array}{cl}
\displaystyle  \min_{y, \sigma_{i,j}} & \mathcal{L}_y(J)\\ 
    \text { s.t.}  & \mathcal{L}_{y}(\leftindex[V]^{b^\prime}c^t_{B_i}) = {\rm coef}(\sigma_{i,0}+\sum_{j=1}^a\sigma_{i,j} \cdot \leftindex[V]^bf^t_{A_j} ),\\
    &\quad\quad\forall\,\,i = 1,2,\ldots,b\\
    & y_{\bf0} =1,\  y\in \mathscr{S}_{\ell}(g)\\ &\sigma_{i,0}\in\Sigma[x]\cap \re[x]_{2k},\\
    & \sigma_{i,j}\in\Sigma[x]\cap \re[x]_{2k - \operatorname{deg}(\leftindex[V]^bf^{t+\Delta t}_{A_j})},\\
    &\quad\quad\forall\,\, i = 1,2,\ldots,b,\,\forall \,\,j=1,2,\dots,a.
\end{array}
\end{equation}

In the above, $y$ is the tms generated by applying the Riesz functional on all monomials in $u = (\hat{\omega},\hat{v})$,  and $\ell$ is the relaxation order such that $\ell\ge\ell_0:=\lceil\max\{\deg(\leftindex[V]^{b^\prime}c^t_{B})/2,\deg(g)/2,\deg(J)/2\}\rceil$. One should recognize that the optimization problem (\ref{eq:mom_relx}) is now an SDP, which seeks to find SOS multipliers $\sigma_{i,j}$ and tms vector $y$ such that the linear combination of $y$ in $J$ is minimized. We now present a theorem demonstrating the quality for the solution of (\ref{eq:mom_relx}). 
\begin{theorem}
\label{tm:moment}
Let $J_{k,\ell}$ be the minimum value of (\ref{eq:mom_relx}), the following holds:
\begin{enumerate}
\item  The sequence $\{J_{k,\ell}\}$ is monotonically increasing in $\ell$,
and $J_{k,\ell}\le J_k$ for every $\ell$.

\item If $y^{*}$ is the minimizer of (\ref{eq:mom_relx}) and $\mbox{rank} (M_{t}[y^{*}]) = \mbox{rank}( M_{t+1}[y^{*}])=1$ for some $\ell_0\le t<\ell$,
then $u^*:=(\mathcal{L}_{y^*}(u_1),\mathcal{L}_{y^*}(u_2),\ldots,\mathcal{L}_{y^*}(u_6))$ is a safe control that minimizes (\ref{eq:add_safety_condition_qm}).
\end{enumerate}
\end{theorem}

\noindent\textit{Proof.} 1) Since the projection of any tms to $\mathscr{S}_{\ell+1}(g)$ is in $\mathscr{S}_{\ell}(g)$ for all $\ell$,
$J_{k,\ell}$ is monotonically increasing in $\ell$.
For a fixed $\ell$, if $u=(\hat{\omega},\hat{v})$ is feasible for (\ref{eq:add_safety_condition_qm}),
then $[u]_{2\ell}$ is a feasible point of (\ref{eq:mom_relx}).
So $J_{k,\ell}\le J(u)$ for all $u$ that is feasible for (\ref{eq:add_safety_condition_qm}), which implies that $J_{k,\ell}\le J_k$.

2) If $\mbox{rank} (M_{\ell}[y^{*}])=1$ holds at the minimizer $y^*$ of (\ref{eq:mom_relx}), then the tms $y^* = [u^*]_{2\ell}$ by \cite[Theorem~2.7.7]{nie2023moment} with $u^*$ the unique minimizer extracted from $y^*$. Since $u^*$ is a feasible point for (\ref{eq:add_safety_condition_qm}), $J(u^*) \ge J_k$.
Also, we have $ J(u^*)=\mathcal{L}_{y^*}(J) = J_{k,\ell} \le J_k$.
This leads to $J(u^*) = J_k$, i.e., $u^*$ is a minimizer of (\ref{eq:add_safety_condition_qm}). \hfill{\qed}

In Theorem~\ref{tm:moment}, the rank condition with $\mbox{rank} (M_{\ell}[y^{*}])= \mbox{rank} (M_{\ell+1}[y^{*}])$ is called \textit{\textbf{flat truncation}} \cite{nie2013certifying}.
If flat truncation fails to hold for any $t$ such that $\ell_0\le t\le \ell$, then the moment relaxation may not be exact, and we may increase the relaxation order $\ell$.
Usually, the flat truncation will be satisfied after several rounds of updates for $\ell$.
On the other hand, if flat truncation holds with $r:=\mbox{rank} (M_{\ell}[y^{*}])>1$, then we may still extract $r$ controls $u^1,u^2,\dots,u^r\in G$ from $y^*$ such that $J(u^j)=\mathcal{L}_{y^*}(J)$ for all $j=1,2,\dots,r$, and $u^j$ is a minimizer of (\ref{eq:add_safety_condition_qm}) if it is feasible.
It is worthwhile to highlight that we may rarely encounter this situation in practice if we properly set our objective function to converge to one exact global minimizer. However, even though it happens, one can still seek to find a sub-optimal solution by normalizing $u^*:=(\mathcal{L}_{y^*}(u_1),\mathcal{L}_{y^*}(u_2),\ldots,\mathcal{L}_{y^*}(u_6))$ such that it is a feasible point in $G$, denoted as $u^{*}_{\rm norm}$. Then the linear velocity of $u^{*}_{\rm norm}$ is scaled via a standard SOS programming to satisfy $c_{B_i}(u^{*}_{\rm norm})^\top[x]_{d_i}\in \qm[\leftindex[V]^{b^\prime}f^{t+\Delta t}_{A}]_{2k} $, such that a safe control can be obtained. 

\section{Results}
In this section, we provide a comprehensive evaluation of our proposed controller through both simulations and real-world experiments. Firstly, we validate our algorithm on robots modeled with different geometries in an L-shape turn. Next, we demonstrate the effectiveness and advantages of our algorithm for a traversing task in an 8 m$\times$36 m forest environment, where key metrics are recorded and analyzed in comparison with other baseline methods. Finally, we implement our proposed method on the Unitree A1 robot in a cluttered indoor environment to demonstrate its practicality on a real robotic platform. As demonstrated in Fig. \ref{fig:overview}, our proposed local reactive controller is integrated with a path planning module to achieve safe navigation in unknown and cluttered environments. The global path is simply set to a straight line connecting start and goal configuration assuming the environment is completely unknown. The local waypoints are then extracted and tracked by the robot in real time. We keep checking whether the local path is heading towards a potential obstacle area based on sensor data. Once a potential collision is detected, the local path replanning mechanism is triggered and a new feasible reference is searched locally within the sensor range using Jump Point Search (JPS)  algorithm\cite{tordesillas2021faster}. The local free regions are extracted using the convex decomposition algorithm in \cite{7839930}.


\subsection{Simulations}
\subsubsection{L-Shape Turn}
\begin{figure}[t]	
	\centering
	\includegraphics[trim=1cm 0.3cm 0 0.6cm, clip,width=1.2\linewidth]{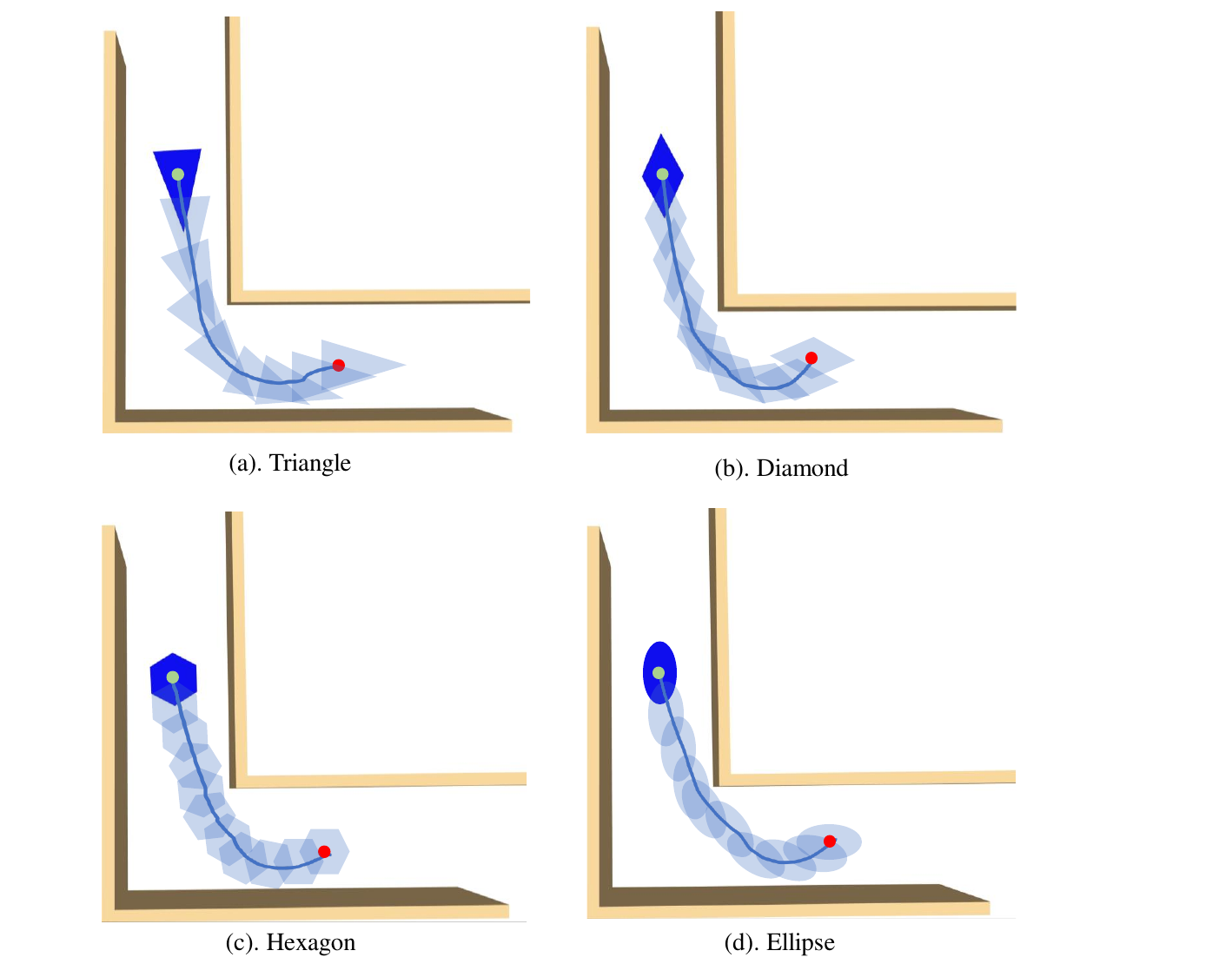}
	\setlength{\abovecaptionskip}{-10pt} 
	\caption
	{Visualization of maneuvers for robots with different geometries in the ``L-turn" simulation. }
	\label{fig:L-turn}
\end{figure}

In the L-shape turn simulation environment shown in Fig. \ref{fig:L-turn}, the robot needs to adapt its pose to make a safe turn at the corner. We test the performance of our proposed controller with four different robot geometries. Nominal twist command is calculated from the error between the current and goal configuration of the robot directly without any local reference in this test and trajectories are visualized. The robots with all shapes are shown to be able to keep seeking chances to move towards the goal configuration directly while avoiding collisions by solving the proposed optimization problem (\ref{eq:mom_relx}). To guarantee safety for various robot geometries, we only need to model the robot geometry with different polynomial inequalities and modify the safety constraints accordingly. Also, we evaluate and record the time needed for solving (\ref{eq:mom_relx}) in Table \ref{tab:L-turn}. Note that $\leftindex[V]^{b^\prime}f^t_{B}(x)$ and $\leftindex[V]^{b^\prime}f^{t+\Delta t}_{A}(x)$ in (\ref{eq:add_safety_condition_qm}) are all SOS-convex polynomials, hence the SOS relaxation at the lowest order is exact \cite{nie2023moment}. Consequently, we find that polytopes with different numbers of facets attain nearly the same average time cost since the order of SOS multipliers in the $\qm$
is set to zero and the SOS multipliers degenerate to nonnegative real numbers. However, in the ellipse case (geometry represented with second-order polynomial inequality), the order of the SOS polynomials is set to 2, which would lead to more variables in the SDP. As a result, relatively more computation time is needed.

\begin{table}[t]
  \centering
  \caption{Optimization Time for Robots with Different Geometries}
    \begin{tabular}{lccc}
    \toprule
    Robot Geometries & Min (ms) & Max (ms) & Avg (ms)\\
    \midrule
    Triangle & 8.71  & 42.66 & 15.36 \\
    Diamond & 10.35 & 32.08 & 16.14 \\
    Hexagon & 10.29 & 63.27 & 16.51 \\
    Ellipse &  10.93 & 73.13 & 22.53 \\
    \bottomrule
    \end{tabular}%
  \label{tab:L-turn}%
\end{table}

\begin{figure*}[t]
    	\centering
        \vspace{3pt}
	\includegraphics[trim=1cm 0cm 0 4cm, clip,width=\linewidth]{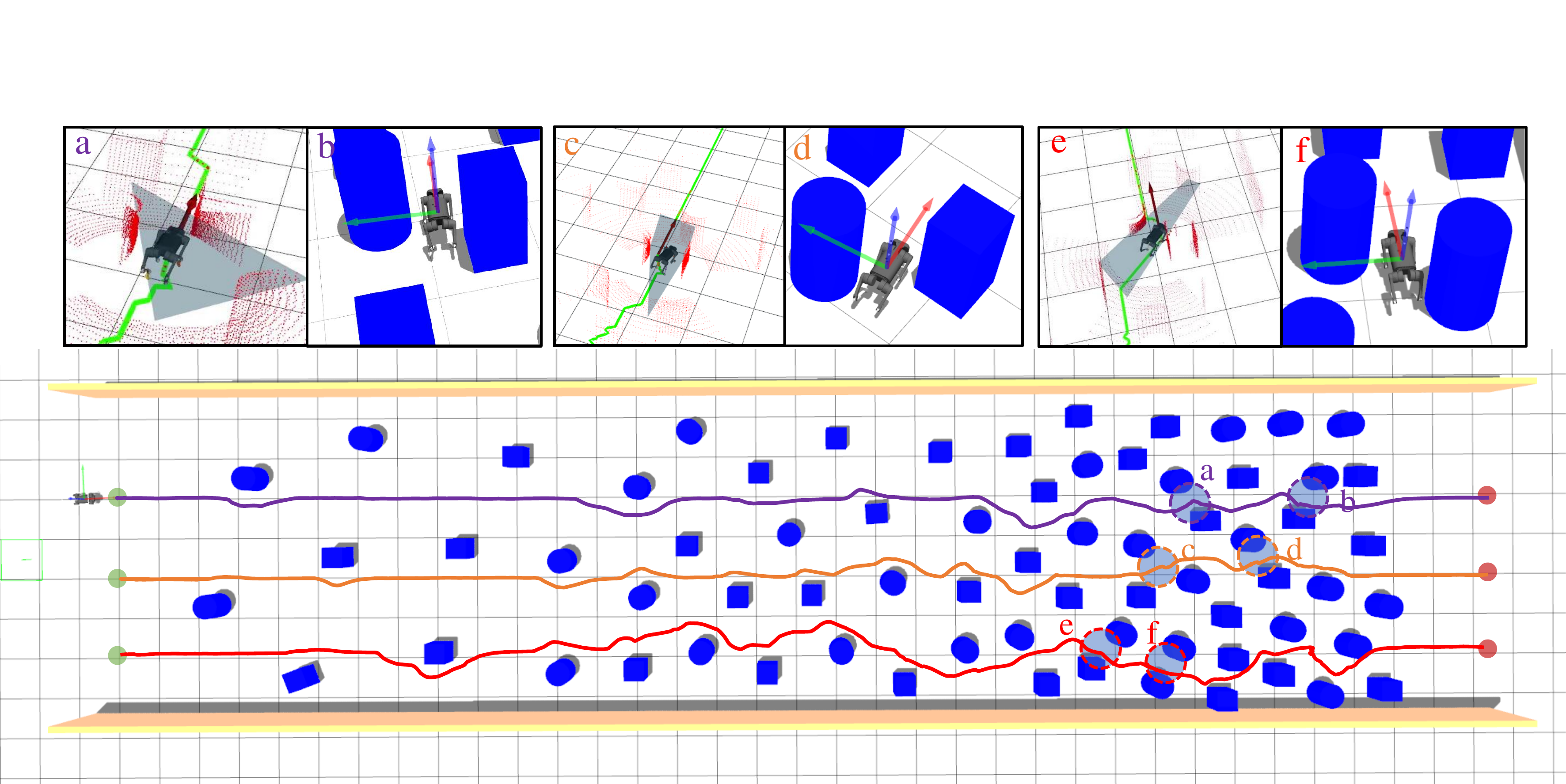}
	\setlength{\abovecaptionskip}{-10pt}
	\caption
	{Trajectories from different start configurations (the green points) generated with our proposed local reactive controller are shown in the forest simulation environment. Key frames are extracted from rviz (a, c, e) and Gazebo (b, d, f) along the trajectories.}
     \label{fig:forest}
\end{figure*}

\subsubsection{Forest}
\label{sec:forest}
In the forest environment that we have built in Gazebo, as depicted in Fig. \ref{fig:forest}, the narrowest passage for the robot to traverse through is 0.4 m wide, and the density of obstacles increases along the path.
We aim to evaluate the performance of our proposed controller on the robot for navigation in unknown cluttered space. In the traversing task, our robot is modeled as a 0.3 m $\times$0.63 m rectangle, which tightly encapsulates its actual geometry. 
The weighting for tracking position and orientation error is set to $Q_p:=diag(5,2,0)$ and $Q_R:=0.1\times \operatorname{I}_{9}$ in this simulation scenario. We choose to compare with the benchmark method in \cite{mattamala2022efficient}, which achieves autonomous safe navigation in challenging environments based on the designed Riemannian Motion Policies (RMP).
We obtain two motion patterns by tuning the weighting between the goal distance field (GDF) and the signed distance field (SDF): aggressive RMP has higher goal confidence to track the reference while conservative RMP shows higher obstacle awareness to keep the robot away from obstacles. The comparison results are obtained by executing each method in the forest environment for 30 times with the same local path guidance.

We denote reaching the final goal configuration as one success traversing. Observed from Table \ref{tab:forest_tab}, the aggressive RMP is able to keep tracking the local reference path but with highest tendency to collide with obstacles, while its conservative counterpart demonstrates safer behaviours. Yet, conservative RMP may hesitate and get stuck at narrow corridors like the ones shown in Fig. \ref{fig:forest}(b), Fig. \ref{fig:forest}(d), and Fig. \ref{fig:forest}(f). We measure the Mean Squared Error (MSE) between actual trajectories and local references of the robot to evaluate pose tracking performance. As indicated in Table \ref{tab:forest_tab}, our controller achieves the lowest pose tracking error while navigating the robot through the cluttered environment, with no collision happening in all repeated simulations. Also, to quantify the efforts for bypassing the obstacles, we define $\eta$ as the length proportion between actual path $D_a$ and the straight line connecting start and goal configurations $D_s$. Different from pushing away the robot under the influence of signed distance field, our method aims to adopt aggressive collision avoidance maneuvers near the obstacles by keeping the robot in the local free region while minimizing the tracking error. 
As a result, our controlled robot attains the lowest $\eta$ value of 1.148.   
\begin{table}[t]
  \centering
  \caption{Comparison of key metrics between our proposed method and RMP \cite{mattamala2022efficient} in the forest environment.}
    \begin{tabular}{crccc}
    \toprule
    \multicolumn{2}{l}{} & Ours  & RMP (Cons.) & RMP (Agg.) \\
    \midrule
    \multicolumn{1}{l}{\multirow{3}[6]{*}{Pose Tracking Error (m)}} & \multicolumn{1}{c}{Min} & 0.117 & 0.225 & 0.249 \\
\cmidrule{2-5}          & \multicolumn{1}{c}{Max} & 0.272 & 0.336 & 0.361 \\
\cmidrule{2-5}          & \multicolumn{1}{c}{Avg} & 0.159 & 0.271 & 0.282 \\
    \midrule
    \multicolumn{1}{l}{\multirow{3}[6]{*}{Collsion Frequency}} & \multicolumn{1}{c}{Min} & 0     & 0     & 0 \\
\cmidrule{2-5}          & \multicolumn{1}{c}{Max} & 0     & 2     & 3 \\
\cmidrule{2-5}          & \multicolumn{1}{c}{Avg} & 0     & 0.75  & 1.27 \\
    \midrule
    Success Rate (\%) &       & 100   & 73    & 100 \\
    \midrule
    $\eta = D_a/D_s$ &       & 1.148 & 1.316 & 1.273 \\
    \bottomrule
    \end{tabular}%
  \label{tab:forest_tab}%
\end{table}%
Our method relies on the extraction of local convex free regions, consuming 0.006 ms in average, which is negligible compared with generating the local map representations (SDF, GDF, etc,.) as illustrated in Table \ref{tab:forest_tab_time}. Besides, it takes around 20 ms to solve for the optimal twist command through the proposed convex conic programming in (\ref{eq:mom_relx}), achieving a control frequency of around 50 Hz.   

\begin{table}[t]
  \centering
  \caption{Comparison of Processing and Traversing Time between our proposed method and RMP\cite{mattamala2022efficient} in the forest environment.}
    \begin{tabular}{p{9.19em}ccc}
    \toprule
    \multicolumn{1}{c}{} & Ours  & \multicolumn{1}{p{5.625em}}{RMP (Cons.)} & \multicolumn{1}{p{5.065em}}{RMP (Agg.)} \\
    \midrule
     Map Processing (ms) & 0.006  & 53.600  & 60.500  \\
    Optimization (ms) & 22.791  & 1.522  & 1.231  \\
    Traversing (s) & 91.007  & 101.136  & 88.040  \\
    \bottomrule
    \end{tabular}%
  \label{tab:forest_tab_time}%
\end{table}%
\subsection{Real-World Experiment}
In the real-world experiments, we implement our proposed local reactive controller on the Unitree A1 quadruped. The perception data is obtained via an onboard Mid-360 Livox LiDAR and the optimal twist command is computed from a GPD Pocket3 with an Intel i7-1195G7 processor. The calculated twist commands are tracked with the whole-body control module equipped on the quadruped. The quadruped is commanded to traverse through an indoor area of 3 m $\times$ 12 m, where obstacles are placed randomly along the path. 

\begin{figure}[t]
	\centering
	\includegraphics[trim=1cm 1cm 1cm 1cm, clip,width=1\linewidth]{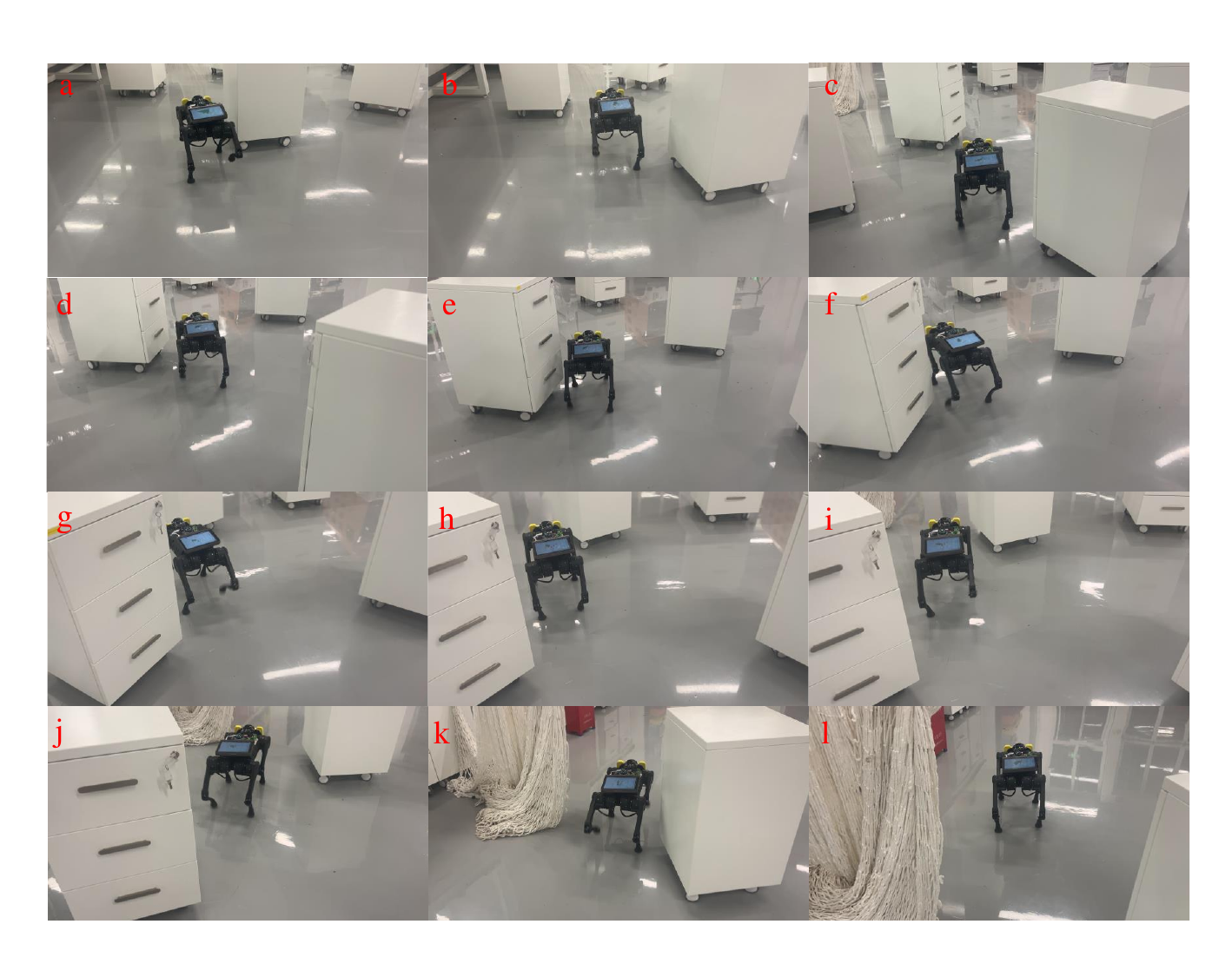}
	\setlength{\abovecaptionskip}{-5pt} 
	\caption
	{Real-world implementation on Unitree A1 quadruped. The robot is commanded to traverse through an unknown indoor space where the obstacles are placed randomly along the path.}
	\label{fig:real_exp}
\end{figure}

As demonstrated in Fig. \ref{fig:real_exp}, the robot is able to traverse through the obstacles based on local perception data. By restricting the robot motion inside the extracted local free regions, our robot avoids all the unknown obstacles in the cluttered environment. The robot attains an average traversing speed of 0.4 m/s along the cluttered path and the computation time for solving the proposed convex conic programming is 21 ms in average. In particular, to obtain different motion patterns, we change the weighting matrices of the objective function $J$ concerning relative importance of tracking position error and orientation error during the traversing task. At the beginning, the robot has higher orientation error awareness, which leads to a tendency for avoiding collisions with translation motion shown in Fig. \ref{fig:real_exp}(a)-\ref{fig:real_exp}(d). Note that the local reference path is close to obstacles because we treat the robot as a point and expect a non-conservative guidance in the path searching step. We increase $Q_p$ and observe from Fig. \ref{fig:real_exp}(e)-\ref{fig:real_exp}(j) that the robot tends to adapt its orientation to bypass the obstacles with a short distance to the obstacles for minimizing the pose tracking error. Overall, our approach demonstrates robust traversing ability in the cluttered environment, and its practicality on real robotic platform is validated.
\section{CONCLUSION}
This work presents an innovative safety-critical local reactive controller that enables robots to navigate in unknown and cluttered environments based on local free regions extracted from onboard sensor data directly. The proposed polynomial optimization framework exhibits the capability to incorporate constraints and objective terms represented as polynomial functions, which can be tailored to specific problem domains. Also, theoretical safety guarantees considering robot geometry are introduced to the optimization framework through the use of polynomial nonnegativity theorems. Additionally, we derive and apply two semidefinite relaxation schemes to the proposed polynomial optimization problem with safety conditions, resulting in an SDP that facilitates real-time reactive motion generation of the robot.  
The proposed controller is evaluated in both numerical simulations and real-world experiments, demonstrating its effectiveness and practicality in generating tight-fitting obstacle avoidance maneuvers in traversing through challenging environments.   
\bibliographystyle{IEEEtran}
\normalem
\bibliography{ref}

\end{document}